\documentclass[12pt,reqno]{amsart}
\usepackage{amsmath,amsfonts,amssymb,amsthm,hyperref,enumerate,multicol}
\usepackage{booktabs}
\usepackage{graphicx}
\usepackage{tikz}
\usetikzlibrary{shapes.geometric, arrows}
\newtheorem{theorem}{Theorem}

\newtheorem{example}{Example}

\numberwithin{equation}{section}
\usepackage{algorithm}
\usepackage{algorithmic}

\begin{document}
\title[]
{A comparative study of some wavelet and sampling operators on various features of an image}
    
\maketitle

\begin{center}
{\bf Digvijay Singh$^1$ Rahul Shukla$^2$, Karunesh Kumar Singh$^3$ } \footnote{Corresponding author: Karunesh Kumar Singh} \\
\vskip0.15in
$^{1,3}$ Department of Applied Sciences and Humanities, Institute of Engineering and Technology, Lucknow, 226021, Uttar Pradesh, India \\
$^{2}$ Department of Mathematics, Deshbandhu College, University of Delhi, India \\
\vskip0.15in

\vskip0.15in

Email: dsiet.singh@gmail.com$^1$, rshukla@db.du.ac.in$^2$,   kksiitr.singh@gmail.com$^3$,
\end{center}

\begin{abstract}
This research includes the study of some positive sampling Kantorovich operators (SK operators) and their convergence properties. A comprehensive analysis of both local and global approximation properties is presented using sampling Kantorovich (SK), Gaussian, Bilateral and the thresholding wavelet-based operators in the framework of SK-operators. Explicitly, we start the article by introducing the basic terminology and state the fundamental theorem of approximation (FTA) by imposing the various required conditions corresponding to the various defined operators. We measure the error and study the other mathematical parameters such as the mean square error (MSE), the speckle index (SI), the speckle suppression index (SSI), the speckle mean preservation index (SMPI), and the equivalent number of looks (ENL) at various levels of resolution parameters. The nature of these operators are demonstrated via an example under ideal conditions in tabulated form at a certain level of samples. Eventually, another numerical example is illustrated to discuss the region of interest (ROI) via SI, SSI and SMPI of 2D Shepp-Logan Phantom taken slice from the 3D image, which gives the justification of the fundamental theorem of approximation (FTA). At the end of the derivation and illustrations we observe that the various operators have their own significance while studying the various features of the image because of the uneven nature of an image (non-ideal condition). Therefore, to some extent, some operators work well and some do not for some specific features of the image. 
\end{abstract}

\section{Introduction }

The approximation theory has broad prospective in pure mathematics. Subsequently, the area of study covers almost every branch of pure and applied mathematics including interpolation of polynomial, sampling operators (SK operators) etc. because of the study of convergence properties. From the application point of view, there are some areas like sampling theory, Wavelet Theory, and Neural Networks in the form of Signal Processing, Machine Learning and Data Science, receptively. One of the beautiful applications of approximation theory comes from the sampling. Basically, in this research we will discuss the various forms of linear and nonlinear sampling Kantorovich operators.  
\par
In the last few years, the evolving research related to this article comes from the Department of Mathematics \& Computer Science, University of Perugia in the direction of SK operators. A considerable amount of scholars \cite{A1,P17, P18, P19, P20} achieved a remarkable landmark in the area of approximation theory. As a result, Bardaro and his colleague discovered a novel SK-operators defined as follows;   

\begin{eqnarray}\label{DSK}
    \mathrm{S}_n(f)(x) := n^d \sum_{k \in \mathbb{Z}^d} \xi(n x - k) \int_{R_k^n} f(u) \, du,
\end{eqnarray}
where the meaning of the parameters $\xi$ is given in Section \eqref{SEC2}. For more detail, see the article viz.\cite{s5,s6,s3,s2,s1,s4,R6}.  \par
In this research, several operators are discussed which come from the deduction of sampling-type \& wavelet-type operators. Gaussian and bilateral-type operators are the deduction of Sk operators, whereas thresholding wavelet-based operators come from wavelet-type operators. The equations \eqref{2dgaussian}, \eqref{2dbilateral} and \eqref{wave} show the mathematical form of the gaussian, bilateral and thresholding wavelet-based operators, receptively;

\begin{eqnarray} \label{2dgaussian}
    (G_\sigma f)(x) = \int_{\mathbb{R}} f(y) \cdot \frac{1}{\sqrt{2\pi} \sigma} \exp\left( -\frac{(x - y)^2}{2\sigma^2} \right) dy.
\end{eqnarray}
In order to study the localization properties in terms of convergence and estimation, the discrete form of the operators is given by
$$
(G_\sigma f)[n] = \sum_{k=-K}^{K} f[n - k] \cdot \frac{1}{\sqrt{2\pi} \sigma} \exp\left( -\frac{k^2}{2\sigma^2} \right),
$$

where $K = \lfloor 3\sigma \rfloor$.

The bilateral operators preserve edges by combining spatial and intensity similarities:
\begin{equation}\label{2dbilateral}
    (B_\sigma f)(x) = \frac{1}{K(x)} \int_{\mathbb{R}} f(y) \cdot \exp\left( -\frac{(x - y)^2}{2\sigma_s^2} \right) \cdot \exp\left( -\frac{(f(x) - f(y))^2}{2\sigma_r^2} \right) dy,
\end{equation}

where $C(x)$ is the normalization constant.
Again, defining the discrete form of the above-defined operators is given as follows;
\begin{equation*}
(B_\sigma f)[n] = \frac{1}{C_n} \sum_{k=-K}^{K} f[n + k] \cdot \exp\left( -\frac{k^2}{2\sigma_s^2} \right) \cdot \exp\left( -\frac{(f[n] - f[n + k])^2}{2\sigma_r^2} \right).
\end{equation*}
For further study, avid readers are advised to see\cite{R4}, in which it has been shown that bilateral operators are faster and more accurate filtering operators. \\
 Let $\psi_{j,k}$ be the wavelet basis functions as mentioned in \cite{R8,R6}. The wavelet operators are defined as follows;
\begin{equation}\label{wave}
    (W_\tau f)(x) = \sum_{j,k} T(w_{j,k}) \cdot \psi_{j,k}(x),
\end{equation}
where $T(\cdot)$ is a thresholding operator (e.g., soft/hard thresholding).\par
In the context of operatorsing techniques, the deduction of SK operators under specific conditions reveals the derivation of Gaussian, Kantorovich, Bilateral and thresholding wavelet-based operators. More preciously, Gaussian and Kantorovich operators directly fit the integral-averaging pattern, while the Bilateral and Wavelet operators require more delectate assumptions involving kernels and projection onto basis functions. 
Estimation and rate of convergence are established using standard continuity arguments, as discussed in the articles \cite{R3, V1, V2, R1, R2, DS, R7}.\par
The novelty of this work lies in the introduction of Gaussian, bilateral, thresholding wavelet-based, and Kantorovich-type operators in the framework of sampling Kantorovich operators and their fundamental theorem approximation discussed in Section $(1,3)$. In Section $(4)$, we investigate the behavior of a smooth 3D test function by applying Gaussian, Bilateral, Wavelet, and Kantorovich-type operators and extend this idea to the discrete nature of functions in terms of ROIs namely 2D Shepp-Logan Phantom. Eventually, we ended up this discussion with the detailed discussion of MSE under regular condition and some other mathematical parameters like SI, SSI, SMPI and ENL for the better applicability of the operators according to the various features 2D Shepp-Logan Phantom image.
\section{Preliminaries;}\label{SEC2}
\subsection{\textbf{Classical SK-operators}}
Let $ f: \mathbb{R}^d \rightarrow \mathbb{R}$ be a locally integrable function. The classical SK- operators are defined as \eqref{DSK},
where \( \xi \) is a kernel that satisfies standard conditions and \( D_a^n \) is the cell:
\[
D_a^n = \prod_{j=1}^{d} \left[\frac{a_j}{n}, \frac{a_j + 1}{n}\right),
\]
where \( \xi: \mathbb{R}^d \to \mathbb{R} \) is a kernel function satisfying the following properties:
\begin{itemize}
    \item[(i)] \textbf{Unit summation property:} \( \sum_{k \in \mathbb{Z}^q} \xi(x - k) = 1 \) for all \( x \in \mathbb{R}^d \),
    \item[(ii)] \textbf{Kernel Boundedness:} \( \sup_{x \in \mathbb{R}^q} |\xi(x)| < \infty \),
    \item[(iii)] \textbf{Kernel estimation:} There exists \( \delta > 0 \) such that \( |\xi(x)| \leq L (1 + \|x\|)^{-q - \delta} \) for some constant \( L > 0 \).
\end{itemize}

\subsection{\textbf{SK-operators in terms of an image}}
    
An image is a representation of the matrix in which each entry denotes the pixel values, but to define the operators in terms of an image using sampling kantorovich operators a 2-dimensional digital grayscale image can be represented
using a step function $ I $ belonging to $L_p(R^2), 1 \leq p < +\infty$, $I$ is defined by:

$$I{(t_1,t_2)}=\sum_{i=1}^{m}\sum_{i=1}^{n} b_{ij}\times \bold{1}_{ij}$$

In the above expression, $\mathbf{1}$ denotes the characteristic function such that if $(t_1, t_2) \in (i, i-1] \times (j, j-1]$, then $\mathbf{1} = 1$; otherwise, its value is $0$.

\subsection*{Arbitrary Setting}

Let \( f : \mathbb{R}^q \to \mathbb{R} \) be a measurable function, and let \( \mathfrak{S}_n(f) \) be its approximation using the SK-operators. Then, define the following parameters:

\subsection{\textbf{Point-wise Error}} 
If $\varsigma_n(x)$ denotes the error between $ \mathfrak{S}_n(f)(x) $ and $f(x)$, then the point-wise error is defined as:
\[
\varsigma_n(x) = \left| \mathfrak{S}_n(f)(x) - f(x) \right|.
\]

\subsection{\textbf{Basic Classical Matrices in Image Form}}

Let \( f(i,j) \in [0,1] \) represent a grayscale image and \( S_n(f)(i,j) \) its deterministic or stochastic approximation.

\subsection{\textbf{Peak Signal-to-Noise Ratio (PSNR)}}
\[
\text{PSNR}(f, \hat{f}) = 10 \cdot \log_{10} \left( \frac{L^2}{\text{MSE}(f, \hat{f})} \right),
\]
where \( L \) is the maximum possible pixel value (e.g., \( L = 1 \) for normalized images), and
\[
\text{MSE}(f, \hat{f}) = \frac{1}{mn} \sum_{i=1}^m \sum_{j=1}^n (f_{i,j} - \hat{f}_{i,j})^2.
\]

\subsection{\textbf{Speckle Index (SI)}} Let us denote \( \sigma_I \) is the standard deviation. In addition, assume that \( \mu_I \) is the mean intensity of the image. Then Speckle Index (SI) is given as follows; 

\[
\mathrm{SI} = \frac{\sigma_I}{\mu_I}
\]

\subsection{\textbf{Speckle suppression index (SSI)}}
The SSI is a mathematical parameter that quantifies the level of noise reduction compared to the original image. Mathematically, it is defined by the following mathematical expression;
\[
\mathrm{SSI} = \frac{\mathrm{SI}_{\text{filtered}}}{\mathrm{SI}_{\text{original}}}
\]
\textbf{Note 1} If the values are closer to 0 it means the image has better speckle suppression.

\subsection{\textbf{Speckle Mean Preservation Index (SMPI)}}
The SMPI evaluates the extent to which the mean intensity is preserved after being operated, which is given by:
\[
\mathrm{SMPI} = \frac{\mu_{\text{filtered}}}{\mu_{\text{original}}}
\]
\textbf{Note 2} Similarly, if SMPI has value close to 1 then it means that the image has excellent mean preservation.

\subsection{\textbf{Equivalent Number of Looks (ENL)}}
To assess the nature of smoothness of an image, we use mathematical parameters defined as;

\[
\mathrm{ENL} = \frac{\mu^2}{\sigma^2}
\]
where \( \mu \) and \( \sigma \) are the mean and standard deviation of intensities in a uniform region.
\textbf{Note 3} Higher values of ENL correspond to better smoother images.

\vspace{0.5cm}

\subsection{\textbf{Thresholding Operator}}
Let \( \lambda > 0 \) be a fixed threshold. A \emph{thresholding operator} \( T_\lambda \) is a nonlinear function that modifies a coefficient \( w \in \mathbb{R} \) such that:
\[
T_\lambda(w) =
\begin{cases}
\phi(w, \lambda), & \text{if } |w| > \lambda \\
0, & \text{if } |w| \leq \lambda
\end{cases}
\]
where \( \phi(w, \lambda) \) is a user-defined function that reduces the magnitude of \( w \), often used to suppress noise or small perturbations in signal or image processing.

\subsection{\textbf{Hard Thresholding}}
Let \( \lambda > 0 \) be a fixed threshold. The \textit{hard thresholding operator} \( T_{\lambda}^{\text{hard}} \) is defined for any coefficient \( w \in \mathbb{R} \) as:
\[
T_{\lambda}^{\text{hard}}(w) =
\begin{cases}
w, & \text{if } |w| > \lambda \\
0, & \text{if } |w| \leq \lambda
\end{cases}
\]
This operator retains only those coefficients whose magnitude exceeds \( \lambda \).

\vspace{0.3cm}

\subsection{\textbf{Soft Thresholding}}
Let \( \lambda > 0 \) be a fixed threshold. The \textit{soft thresholding operator} \( T_{\lambda}^{\text{soft}} \) is defined as:
\[
T_{\lambda}^{\text{soft}}(w) =
\begin{cases}
\text{sgn}(w) \cdot (|w| - \lambda), & \text{if } |w| > \lambda \\
0, & \text{if } |w| \leq \lambda.
\end{cases}
\]

\vspace{0.3cm}

\section{Main results}


We present detailed proofs showing that four commonly used image operatorss can be interpreted as approximation operators and that they converge to the identity operator under certain assumptions.

\begin{theorem}
    Let \( f \in L^p(\mathbb{R}^d) \), \( 1 \leq p < \infty \). Further, assume that the kernel function known as mollifier (normalized, positive, smooth) is defined as follows;
\[
\varphi_\sigma(x) = \frac{1}{(2\pi \sigma^2)^{d/2}} e^{-\|x\|^2/(2\sigma^2)}.
\]
Again, from equation \eqref{2dgaussian}, we have the uni-dimensional version of Gaussian operators are given by the following expression;
\[
(G_\sigma f)(x) = (f * \varphi_\sigma)(x) = \int_{\mathbb{R}^d} f(y) \varphi_\sigma(x - y) \, dy.
\]. Then, 
\[
\lim_{\sigma \to 0} \|G_\sigma f - f\|_{L^p} = 0.
\]
\end{theorem}
 The idea of proof of the above theorem is given in these articles\cite{I1,I2}.

\begin{theorem}
    
\end{theorem}

 Let \( f \in L^p(\mathbb{R}) \), \( 1 \leq p < \infty \). The generalized form of sampling Kantorovich operators is defined as follows;
\[
(S_w f)(x) = \sum_{k \in \mathbb{Z}^d} w^d \left[ \int_{R_k^w} f(u) \, du \right] \varphi(w(x - x_k)),
\]
where \( x_k = \frac{k}{w} \) and the kernel \( \varphi \) satisfies:
\begin{itemize}
    \item \( \int \varphi = 1 \), \( \varphi \in L^1 \cap L^\infty \)
    \item \( \int \|x\|^r |\varphi(x)| dx < \infty \).
\end{itemize}
We have \[
\lim_{w \to \infty} \|S_w f - f\|_{L^p} = 0.
\]
A keen reader can establish the proof of the above theorem with the help of Butzer \& Stens (1991) \cite{I3} and Bardaro et al. (2004)\cite{I4}.

\begin{theorem}
    Given \( f \in \mathrm{Lip}(\mathbb{R}^d) \). 
\[
(B_\sigma f)(x) = \frac{1}{C(x)} \int_{\mathbb{R}^d} f(y) \, G_{\sigma_s}(x - y) \, G_{\sigma_r}(f(x) - f(y)) \, dy,
\]
with normalization constant \( C(x) \). The notation \( G_{\sigma_s} \in L^1 \)
and \( G_{\sigma_r} \in L^\infty \) denote the Gussian and symmetric kernel bounded $\&$, receptively. Then we obtain
\[
\lim_{\sigma \to 0} \|B_\sigma f - f\|_{L^p} = 0,
\]
as \( \sigma_s, \sigma_r \to 0 \), the operatorizing becoming increasingly localized in space and intensity.
\end{theorem}
 Again, the proof can be guessed in the articles \cite{I5,I6} discussed by Tomasi \& Manduchi, 1998; Barash, 2002.

\begin{theorem}
    Let $W^{s,p}(\mathbb{R}^d)$ be a Sobolev space and \( f \in W^{s,p}(\mathbb{R}^d) \). Further assume that \( \{\psi_{j,k}\} \) is a compactly supported orthonormal wavelet basis. From \eqref{wave}, the uni-dimensional wavelet type operators are defined as follows;
\[
(W_J f)(x) = \sum_{j \leq J} \sum_k T(w_{j,k}) \psi_{j,k}(x),
\]
where \( T(\cdot) \) is a thresholding operator. As $J \rightarrow\infty$, we have 
\[
\lim_{J \to \infty} \|W_J f - f\|_{L^p} = 0.
\]

\end{theorem}

\textbf{Proof:}
By Mallat, \cite{I7}; Donoho \& Johnstone, 1994 \cite{I8}; Perrier \& Basdevant, 1996 \cite{I9} and using Jackson-type inequalities:
\[
\|f - W_J f\|_{L^p} \leq C 2^{-Js} \|f\|_{W^{s,p}},
\]
one can derive the proof.



    \section*{Illustrations}

    \begin{example}
    Let $f \in L^2([0,1]^3) $ as defined in\eqref{eq1} which is also a compactly support in $[0,1]^3$. We approximate \eqref{eq1} by Gaussian, Bilateral, wavelets and kantorovich type operators, represented by following equations viz. \eqref{eq2}, \eqref{eq3}, \eqref{eq4} and \eqref{eq5};
\begin{equation}\label{eq1}
    f(x, y, z) = \sin(\pi x)\sin(\pi y)\sin(\pi z), \quad (x, y, z) \in [0, 1]^3.
\end{equation}
Let us denote $G_\sigma(f)(x, y, z)$ be Gaussian Operators, defined as 
\begin{equation}\label{eq2}
    G_\sigma(f)(x, y, z) = \iiint_{\mathbb{R}^3} f(u, v, w) \cdot \frac{1}{(2\pi\sigma^2)^{3/2}} \exp\left(-\frac{(x-u)^2 + (y-v)^2 + (z-w)^2}{2\sigma^2}\right) \, du\,dv\,dw.
\end{equation}
The bilateral operators denoted by $B_\sigma(f)(x, y, z)$ is defined as follows
\begin{equation} \label{eq3}
\begin{aligned}
B_\sigma(f)(x, y, z) = \frac{1}{W(x, y, z)} \iiint_{\Omega} & f(u,v,w) \cdot \exp\left( -\frac{(x-u)^2 + (y-v)^2 + (z-w)^2}{2\sigma_s^2} \right) \\
& \cdot \exp\left( -\frac{(f(x,y,z) - f(u,v,w))^2}{2\sigma_r^2} \right) \, du\,dv\,dw
\end{aligned}
\end{equation}

    where \( W(x, y, z) \) is a normalization factor.
    Further, assume that $W_T(f)$ denote the wavelets type of operators defined as follows
\begin{equation}\label{eq4}
    W_T(f) = \mathcal{W}^{-1} \left( T\left( \mathcal{W}(f) \right) \right)
\end{equation}
    where \( \mathcal{W} \) denotes the wavelet transform, \( \mathcal{W}^{-1} \) is the inverse wavelet transform, and \( T \) is a thresholding operator applied to the wavelet coefficients.
    Let \( f\in L^2([0,1]^3) \) be a 3D function. Its wavelet decomposition at level \( J \) is given by:
\[
\mathcal{W}(f) = \left\{ c_{J}, \{ d_{j}^{(l)} \}_{j=1}^{J},\ l = 1, \dots, 7 \right\}
\]
where \( c_{J} \) represent coefficients at \( J \) level and \( d_{j}^{(l)} \) represent not only coefficients at level \( j \) but also orientation \( l \in \{HHH, HHL, HLH, LHH, LLH, LHL, HLL\} \).

We apply a thresholding operator \( T_\lambda \) (either soft or hard) to the detail coefficients:
\[
T_\lambda(d) =
\begin{cases}
\text{sgn}(d)\cdot(|d| - \lambda), & \text{if } |d| > \lambda \quad (\text{soft thresholding}) \\
0, & \text{if } |d| \leq \lambda \\
d, & \text{otherwise (hard thresholding)}
\end{cases}
\]

The thresholded wavelet coefficients are:
\[
\tilde{d}_{j}^{(l)} = T_\lambda(d_{j}^{(l)})
\]

The denoised function is then reconstructed using the inverse wavelet transform:
\[
W_T(f)(x, y, z) = \mathcal{W}^{-1} \left( c_J, \{ \tilde{d}_{j}^{(l)} \} \right)
\]

where \( \mathcal{W}^{-1} \) is the inverse 3D wavelet transform.
In this discussion, the last operators called `sampling kantorovich operators' denoted by $ K_n(f)(x, y, z)$, given by the following expression

\vspace{1em}
    \begin{equation}\label{eq5}
    K_n(f)(x, y, z) = \sum_{i=0}^{n} \sum_{j=0}^{n} \sum_{k=0}^{n} f\left( \frac{i}{n}, \frac{j}{n}, \frac{k}{n} \right) \cdot \phi_{n}(x - \tfrac{i}{n}) \cdot \phi_{n}(y - \tfrac{j}{n}) \cdot \phi_{n}(z - \tfrac{k}{n}),
\end{equation}
    where \( \phi_n \) is a suitable kernel function (e.g., B-spline or box function).
    Now, we also evaluate the mean square error (MSE) defined as follows 

Given a numerical approximation \( \hat{f} \), the error is computed using:
\[
\mathrm{MSE}(f, \hat{f}) = \frac{1}{N^3} \sum_{i=1}^{N} \sum_{j=1}^{N} \sum_{k=1}^{N} \left( f(x_i, y_j, z_k) - \hat{f}(x_i, y_j, z_k) \right)^2
\]
where \( N \) is the resolution along each axis.


\begin{figure}[htbp]
    \centering
    
        \includegraphics[width=0.8\textwidth]{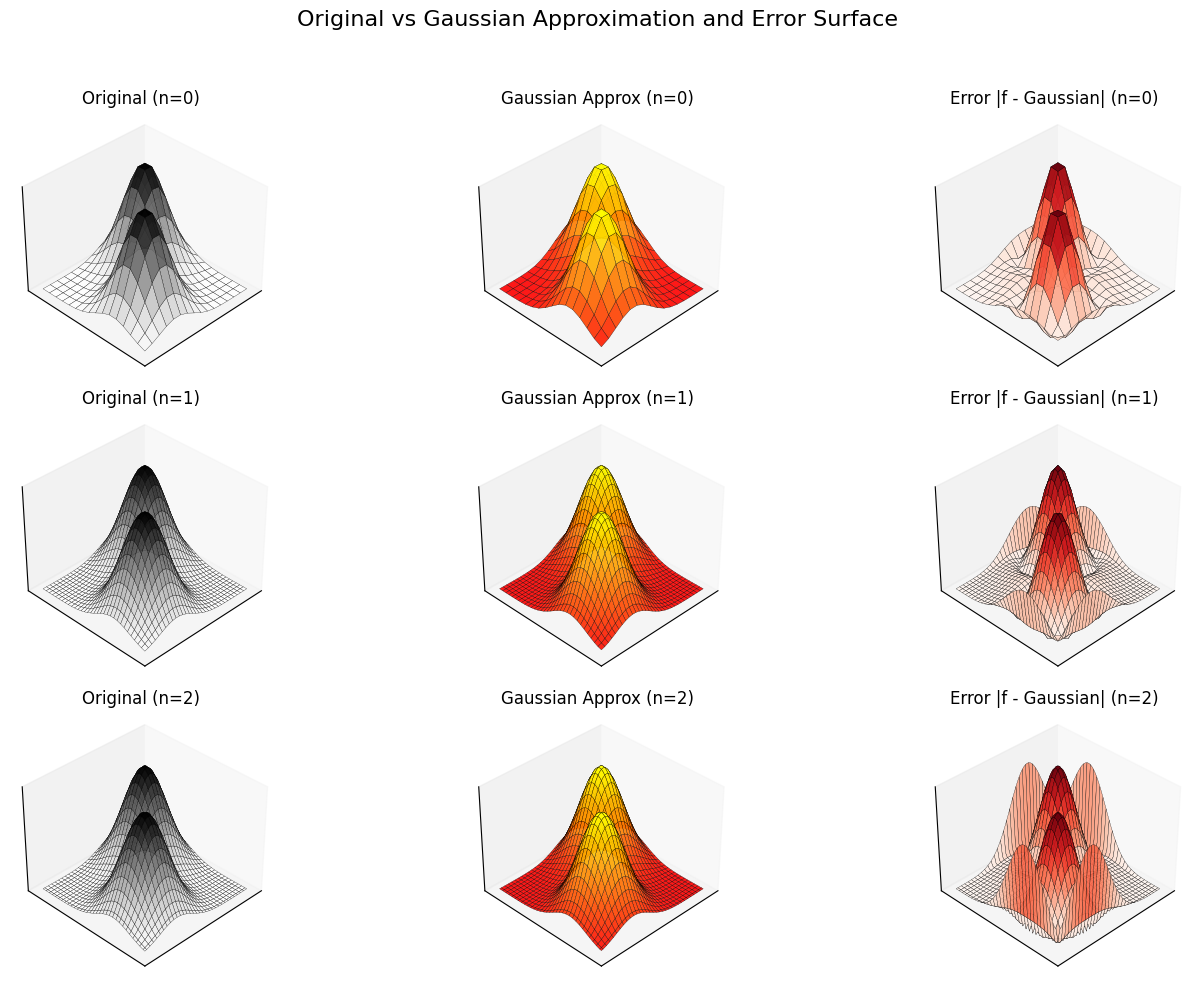}
        \caption{Gaussian Approximation}
    \end{figure}
   \begin{figure}[htbp]
    \centering
        \includegraphics[width=0.8\textwidth]{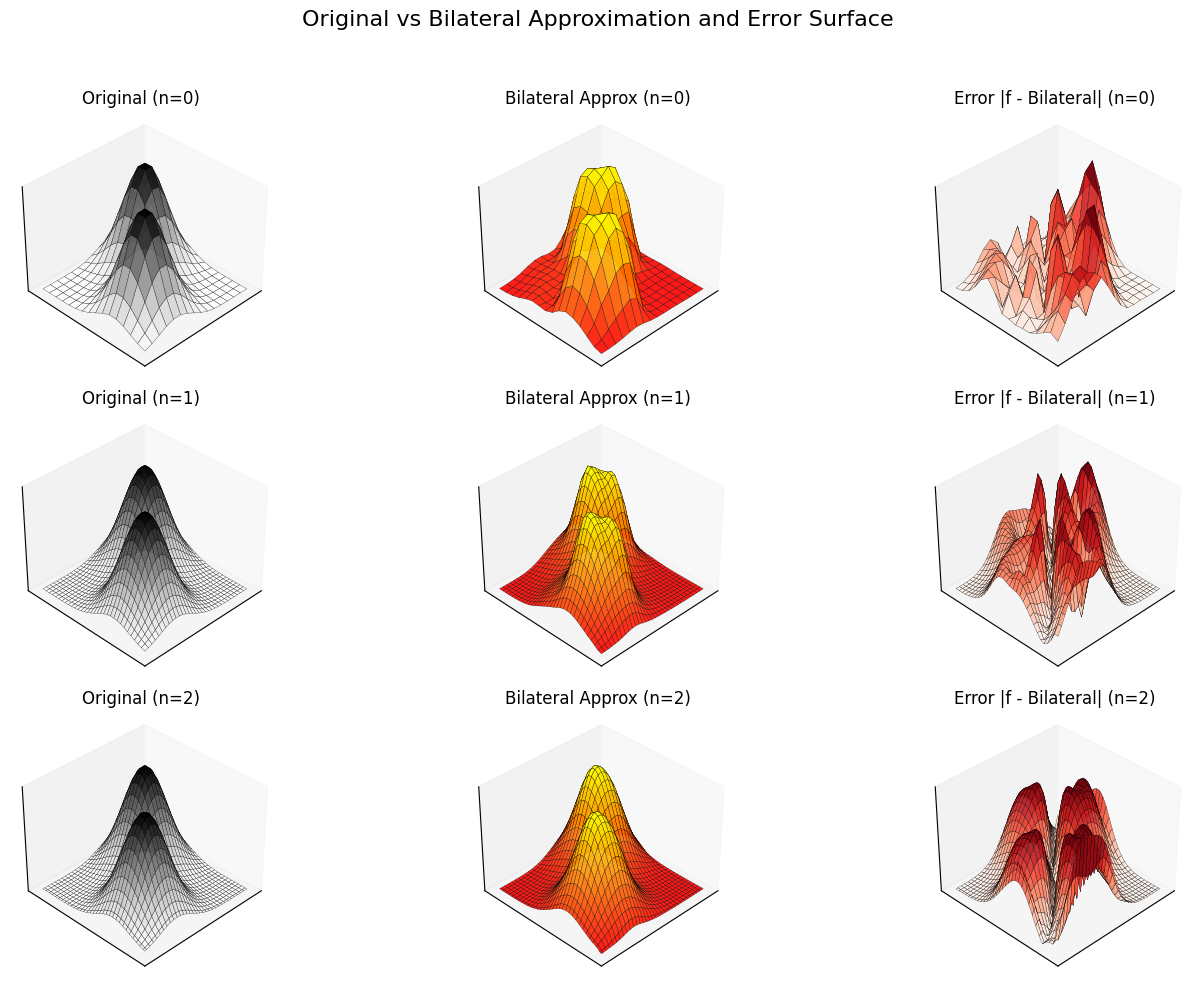}
        \caption{Bilateral Approximation}
    \end{figure}
    \begin{figure}[htbp]
    \centering
        \includegraphics[width=0.8\textwidth]{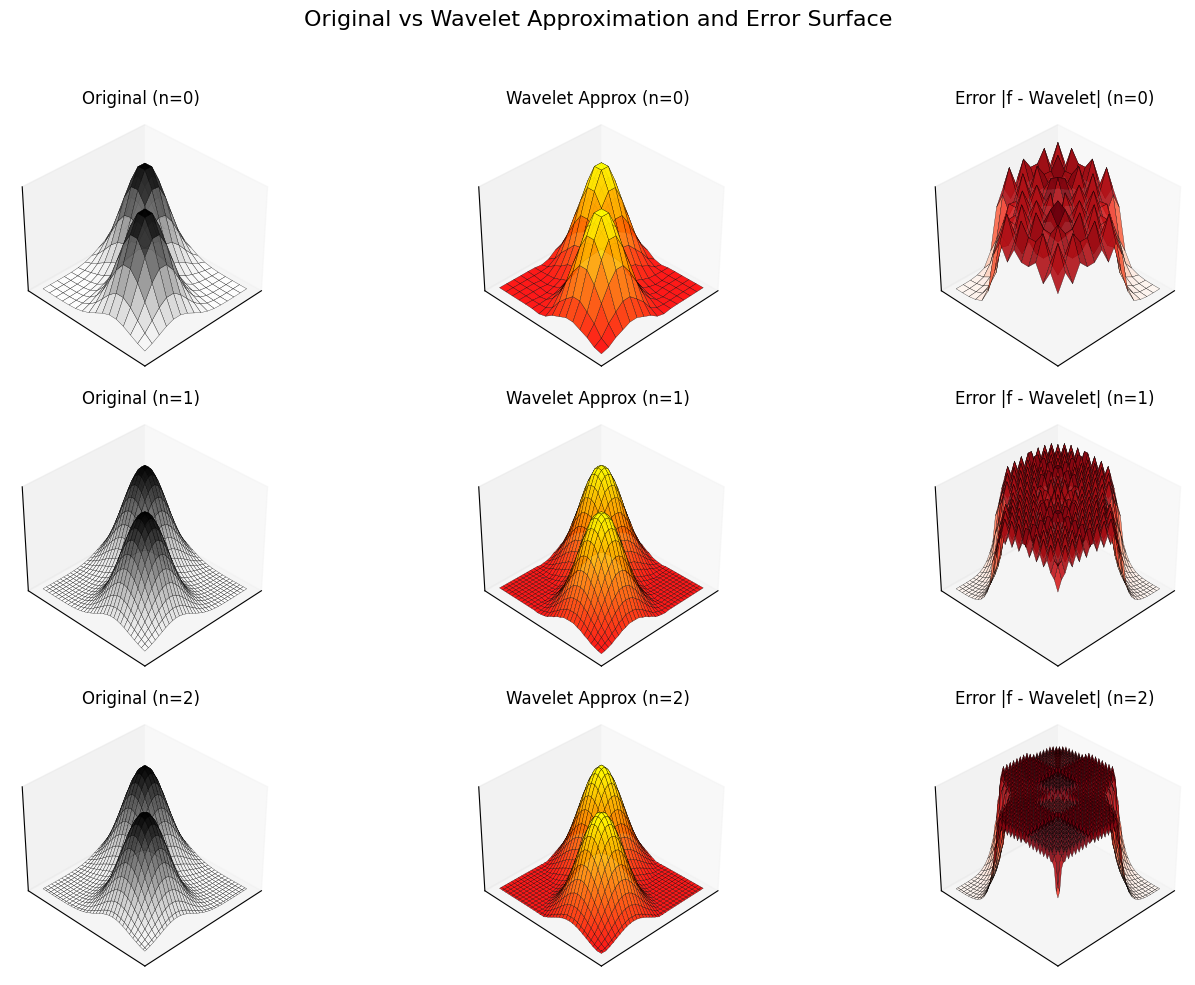}
        \caption{Wavelet Approximation}
\end{figure}
\begin{figure}[htbp]
    \centering
        \includegraphics[width=0.8\textwidth]{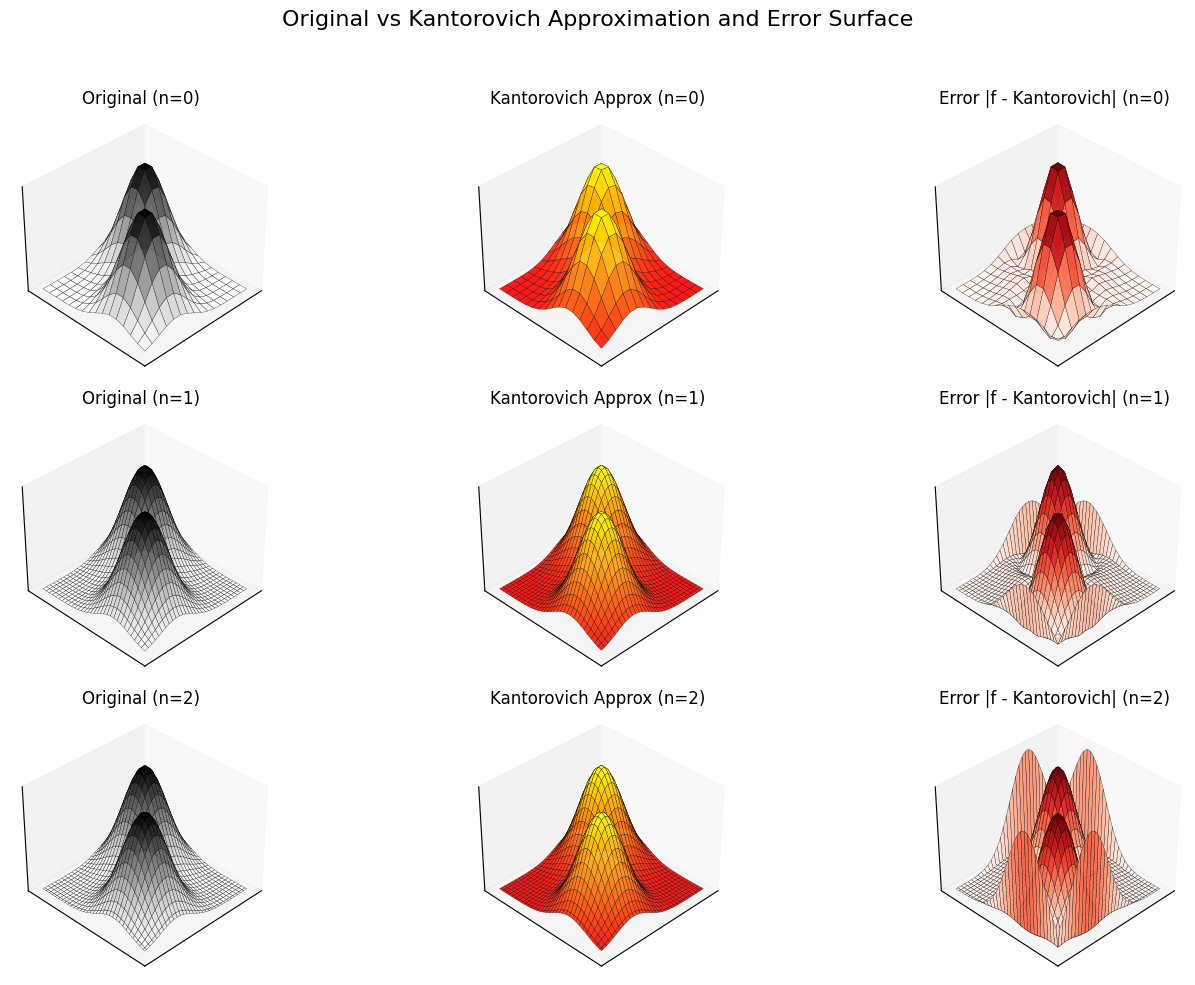}
        \caption{Kantorovich Approximation}

    \caption{3D surface plots of the middle $z$-slice for different approximation operators at varying resolutions ($n = 0,1,2$). Each figure displays the structure-preserving or smoothing effects across resolutions.}
    \label{fig:3d_comparisons}
\end{figure}
\begin{table}[htbp]\label{MSEt}
\centering

\begin{tabular}{|c|c|c|c|c|}
\hline
\textbf{Resolution (n)} & \textbf{Gaussian} & \textbf{Bilateral} & \textbf{Wavelet} & \textbf{Kantorovich} \\
\hline
0 ($16^3$) & 0.02842 & 0.01831 & 0.03785 & 0.02377 \\
\hline
1 ($32^3$) & 0.01495 & 0.00870 & 0.02142 & 0.01109 \\
\hline
2 ($64^3$) & 0.00758 & 0.00392 & 0.01076 & 0.00541 \\
\hline
\end{tabular}
\vspace{0.5cm}

\caption{Mean Squared Error (MSE) between original function and approximated volumes using various operators}
\label{tab:mse-numerical}
\end{table}
\end{example}

\vspace{0.3cm}


\newpage
As observed in Figure~\ref{fig:3d_comparisons}, the bilateral operators clearly demonstrates superior structure preservation and approximation quality at all resolutions. The quantitative errors shown in Table\ref{tab:mse-numerical} reinforce this observation, where bilateral consistently yields the lowest MSE. While Gaussian operatorsing improves gradually with resolution, it fails to preserve edges. Wavelet approximations are unstable due to thresholding artefacts, and the Kantorovich operator, though stable, lacks adaptive sharpness. Table~\ref{MSEt} summarizes these visual trends qualitatively across resolutions.


\newpage
\begin{example}
We assume to simulate a $3-$dimensional volume using the Shepp-Logan phantom and do the investigation in region of interest (ROI) with respect to the mathematical parameters such as structural similarity index (SSI), signal-to-mean-plus-index (SMPI)
and equivalent number of looks (ENL) on the basis of the approximation. In particular, we have identified the white matter (WM), tumor ROI, cerebrospinal fluid (CSF), liver parenchyma, kidney edge, aorta as ROI taken from the mid-slice
of a 3D Shepp-Logan phantom volume. For this ROI, we do the following setting
\begin{itemize}
\item Take the volume: \( \texttt{volume[32,:,:]} \) and select the mid-slice.
    \item Let us specify the ROI with following dimension as;
    
    \begin{enumerate}
        \item White Matter (WM): \texttt{slice\{(50,70) $\times$ (50,70)\}}
        \item Tumor ROI: \texttt{slice\{(30,50)$\times$ (80,100)\}}
        \item CSF: \texttt{slice\{(10,30) $\times$(10,30)\}}
        \item Liver Parenchyma: \texttt{slice\{(70,90)$\times$ (30,50)\}}
        \item Kidney Edge: \texttt{slice\{(80,100)$\times$ (90,110)\}}
        \item Aorta: \texttt{slice\{(40,60)$\times$ (10,30)\}}
    \end{enumerate}
\end{itemize}
We now provide a pseudocode for the proposed problem, which has been implemented in Python. The pseudocode includes the key steps in solving the problems. 

\vskip0.15in
\newpage

\begin{center}
\begin{tabular}{c}
\hline
\textbf{Algorithm:} Pseudocode for the implementation of the proposed problem in Python.\\
\hline
\end{tabular}
\end{center}

\subsection*{Step 1: Volume Generation}

\textbf{Input:} 2D Shepp-Logan Phantom

\textbf{Output:} Synthetic 3D Volume of shape $64 \times 128 \times 128$

\textbf{For} each slice in range $[0, 63]$, \textbf{do:}
\begin{itemize}
    \item Load 2D phantom using \texttt{skimage.data.shepp\_logan\_phantom}.
    \item Resize to $128 \times 128$ using bicubic interpolation.
    \item Stack the resized 2D image along the depth axis to construct a 3D volume.
\end{itemize}

\subsection*{Step 2: Operators implementations}
\textbf{Input:} Extracted RoIs from Step 2 \\
\textbf{Output:} operatorsed RoIs using 4 different techniques \\
\textbf{For} each RoI, apply:
\begin{itemize}
    \item \textbf{Gaussian operators:} Smooth with $\sigma = 1$.
    \item \textbf{Kantorovich operators:} Approximate by applying Gaussian blur iteratively (3 iterations).
    \item \textbf{Bilateral operators:} Use \texttt{sigma\_color}=0.05, \texttt{sigma\_spatial}=1.
    \item \textbf{Wavelet Denoising:} Decompose and threshold using wavelet transform.
\end{itemize}
\noindent\rule{\linewidth}{0.8pt}  

\vskip0.15in
\vskip0.15in


\begin{frame}
	
	\begin{figure}
		\includegraphics[width=0.65\linewidth]{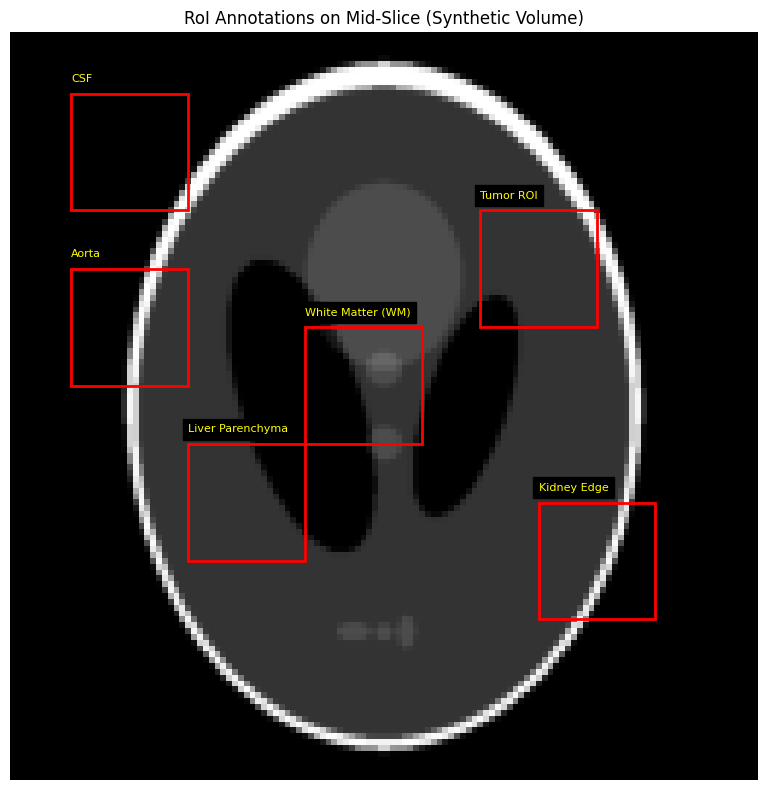}
		\caption{Allocated ROI in 2D Shepp-Logan Phantom }
	\end{figure}
\end{frame}


\vskip0.11in
Moreover, in order to have a precise interference, the parameters like SI, SSI, SMPI, ENL are evaluated and the quantitative evaluation across diverse anatomical and pathological RoIs, the following observations can be drawn:
\vskip0.15in

\begin{table}[h]\label{ROItable}
\centering

\begin{tabular}{|l|l|c|c|c|c|}
\hline
\textbf{ROI Name} & \textbf{Operators} & \textbf{SI} & \textbf{SSI} & \textbf{SMPI} & \textbf{ENL} \\
\hline
White Matter (WM) & Gaussian     & 0.02 & 0.94 & 1.14 & 1.30 \\
                  & Kantorovich  & 0.02 & 0.82 & 1.21 & 1.46 \\
                  & Bilateral    & 0.03 & 0.92 & 1.05 & 1.09 \\
                  & Wavelet      & 0.03 & 1.00 & 1.09 & 1.19 \\
\hline
Tumor ROI         & Gaussian     & 0.03 & 0.97 & 1.94 & 3.78 \\
                  & Kantorovich  & 0.02 & 0.92 & 2.26 & 5.09 \\
                  & Bilateral    & 0.06 & 0.99 & 1.67 & 2.78 \\
                  & Wavelet      & 0.06 & 1.00 & 1.69 & 2.87 \\
\hline
CSF               & Gaussian     & 0.03 & 0.98 & 0.14 & 0.02 \\
                  & Kantorovich  & 0.02 & 0.95 & 0.19 & 0.04 \\
                  & Bilateral    & 0.06 & 1.00 & 0.10 & 0.01 \\
                  & Wavelet      & 0.05 & 1.00 & 0.11 & 0.01 \\
\hline
Liver Parenchyma  & Gaussian     & 0.01 & 0.97 & 2.26 & 5.11 \\
                  & Kantorovich  & 0.01 & 0.93 & 2.42 & 5.84 \\
                  & Bilateral    & 0.02 & 0.98 & 2.20 & 4.83 \\
                  & Wavelet      & 0.02 & 1.00 & 2.13 & 4.54 \\
\hline
Kidney Edge       & Gaussian     & 0.06 & 0.80 & 1.27 & 1.63 \\
                  & Kantorovich  & 0.03 & 0.56 & 1.66 & 2.76 \\
                  & Bilateral    & 0.13 & 0.99 & 0.89 & 0.79 \\
                  & Wavelet      & 0.12 & 1.00 & 0.90 & 0.81 \\
\hline
Aorta             & Gaussian     & 0.07 & 0.80 & 0.81 & 0.65 \\
                  & Kantorovich  & 0.04 & 0.53 & 0.99 & 0.98 \\
                  & Bilateral    & 0.13 & 0.99 & 0.62 & 0.39 \\
                  & Wavelet      & 0.13 & 1.00 & 0.63 & 0.40 \\
\hline

\end{tabular}
\vskip0.15in

\vskip0.15in
\caption{Denoising Metric Results across RoIs and operatorss}
\end{table}
Now, we analyze the various characteristic of an image such as resolution, edge detection etc. of the specified ROIs by employing significant qualitative metrics including SI, SSI, SMPI, and ENL. The following picture \eqref{fig2} includes the sub-pictures, each highlighting distinct ROI. These visual cues offer an empirical justification for the quantitative results presented earlier. By examining these sub-pictures, one can better understand how a method or metric performs, responds, or varies under different conditions.\par
\vskip0.15in

We initiate the investigation by picking a specific sub-picture as an original image and implement it over various aspects such as Gaussian, Kantorovich, bilateral and wavelet operators. As we can see that the same idea can be implemented to the given ROIs viz. white matter, tumor, erebrospinal fluid (CSF), liver parenchyma, kidney edge and aorta corresponding to the Gaussian, Kantorovich, Bilateral and wavelet operators, respectively. The figure \eqref{fig2} depicts the whole senior of the discussion in the next page.

	\begin{figure}\label{fig2}
		\includegraphics[width=1\linewidth]{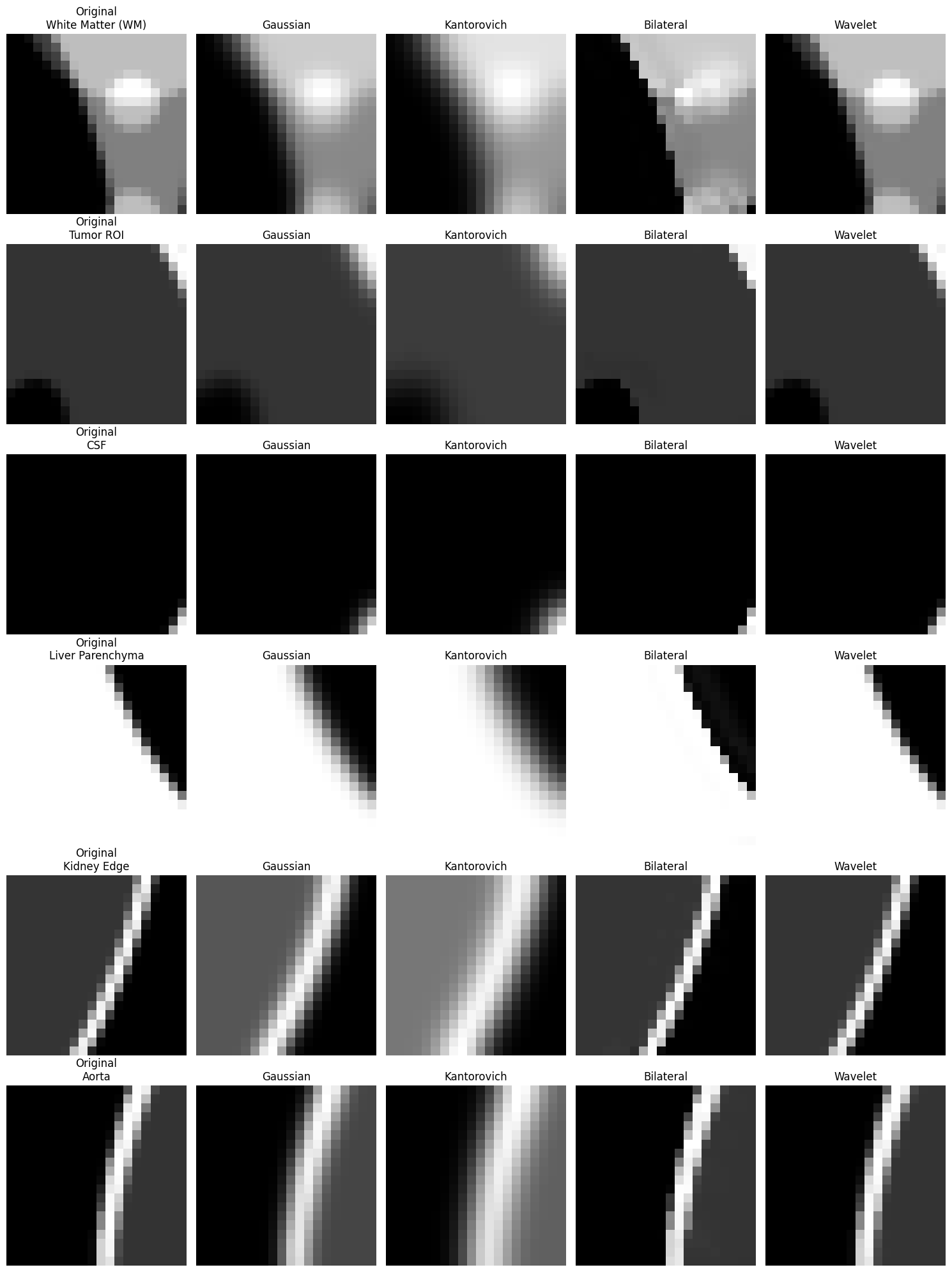}
		\caption{Behavior of ROIs of 2D Shepp-Logan Phantom with respect to the defined Gaussian, Kantorovich, Bilateral and wavelets operators, respectively.  }
	\end{figure}
\newpage
The table\eqref{ROItable} shows that the Kantorovich-type operators have high SMPI and high ENL which means these type of operators consistently provide the robust performance in preserving the edge information filters the unwanted noise. Thresholding wavelet-based operators shine in preserving the SI because of their value in SSI which is useful in structure retention. Bilateral and Gaussian operators perform better and outperformed in most metrics by Kantorovich and thresholding wavelet-based operators.

\end{example}

\section*{Conclusion}
In this research, a significant effort is made to check the behavior of prominent operators such as Gaussian, bilateral, thresholding wavelet-based \& kantorovich type operators. The idea of establishing the fundamental theorem of approximation for these operators is referred to several articles, which set up the base to study their application properties in the form of 3D surface plots of the middle z-slice. We see the nature of approximation in terms of 3D surface plots and Mean Squared Error (MSE) metrics. Furthermore, some ROIs are specified in 2D Shepp-Logan Phantom image and study their approximation properties in the form of SI, SSI, SMPI and ENL. The results show that the bilateral operators consistently have been the best among all operators not only in the ideal cases but also non-idea cases. Explicitly, Bilateral operators preserve the edge structures while suppressing noise which tends to obey more closely with the principles of the Fundamental Theorem of Approximation. In contrast, the Gaussian operators provide smoothing that leads to a loss of important edge detection features. The Kantorovich operators give a stable and uniform approximation but unable to give suitable variations locally. These insights collectively support a comprehensive study of 3D approximation methods with some specific ROIs in which experimental results support the predictions and highlight the effectiveness and limitations of each operators.

\end{document}